\documentclass[%
 reprint,
 amsmath,amssymb,
 aps,
prx,
]{revtex4-1}
\usepackage{xcolor}
\usepackage[draft,nomargin,inline,author=]{fixme}
\fxsetup{theme=color}
\definecolor{fxwarning}{rgb}{0.8,0.0000,0.0000}
\RequirePackage{footnote}
\usepackage{graphicx}
\usepackage{dcolumn}
\usepackage{bm}

\begin{document}

\preprint{APS/123-QED}

\title{Takens-inspired neuromorphic processor: A downsizing tool for random recurrent neural networks via feature extraction}

\author{Bicky A. Marquez$^{1}$}
\email{bama@queensu.ca}
\author{Jose Suarez-Vargas$^{2,3}$}
\author{Bhavin J. Shastri$^{1}$}
\affiliation{%
1. Department of Physics, Engineering Physics \& Astronomy,  Queen's University,  Kingston, Ontario K7L 3N6, Canada. \\
2. Elettra-Sincrotrone Trieste, Strada Statale 14-km 163,5, 34149 Basovizza, Trieste, Italy \\
3. International Centre for Theoretical Physics, Strada Costiera 11, I-34151 Trieste, Italy. \\}

\date{\today}

\begin{abstract}
We describe a new technique which minimizes the amount of neurons in the hidden layer of a random recurrent neural network (rRNN) for time series prediction. 
Merging Takens-based attractor reconstruction methods with machine learning, we identify a mechanism for feature extraction that can be leveraged to lower the network size. 
We obtain criteria specific to the particular prediction task and derive the scaling law of the prediction error. 
The consequences of our theory are demonstrated by designing a Takens-inspired hybrid processor, which extends a rRNN with a priori designed delay external memory. 
Our hybrid architecture is therefore designed including both, real and virtual nodes.
Via this symbiosis, we show performance of the hybrid processor by stabilizing an arrhythmic neural model. 
Thanks to our obtained design rules, we can reduce the stabilizing neural network's size by a factor of 15 with respect to a standard system.
\begin{description}
\item[PACS numbers]
May be entered using the \verb+\pacs{#1}+ command.
\end{description}
\end{abstract}

\pacs{Valid PACS appear here}
\maketitle


\section{Introduction}
Artificial neural networks (ANNs) are systems prominently used in computational science as well as investigations of biological neural systems. In biology, of particular interest are recurrent neural networks (RNNs), whose structure can be compared among others with nervous system's networks of advanced biological species \cite{Maass2016}. In computation, RNNs have been used to solve highly complex tasks which pose problems to other classical computational approaches \cite{Jaeger, Graves2009, Graves2013, Sak2014, Li2014}. 
Their recurrent architecture allows the generation of internal dynamics, and consequently RNNs can be studied utilizing principles of dynamical systems theory. Therefore, the network's nonlinear dynamical properties are of major importance to its information processing capacity. In fact, optimal computational performances are often achieved in a stable equilibrium network's state, yet near criticality \cite{Langton, Natschlager}.  

Among the recurrent networks reported in current literature, rRNNs are popular models for investigating fundamental principles of information processing. In these models the synaptic neural links are randomly weighted, typically following a uniform \cite{Marquez2018} or Gaussian distribution \cite{Bruckstein, Ganguli, Babadi, Schottdorf}. Recently, there is an increasing interest in some particular types of random recurrent networks with a simplified design, where just the output layers are trained using a supervised learning rule. 
Such rRNNs are typically referred to as Reservoir Computers, which include echo state networks \cite{Jaeger} and also liquid state machines \cite{Maass}. Reservoir Computing provides state-of-the-art performance for challenging problems like high-quality long term prediction of chaotic signals \cite{Jaeger, Antonik2017}.

Prediction corresponds to estimating the future developments of a system based on knowledge about its past. Chaotic time series prediction is of importance to a large variety of fields, including the forecasting of weather \cite{lorenz}, the evolution of some human pathologies \cite{Mackey}, population density growth \cite{ottbook}, or dynamical control as found in the regulation of chaotic physiological functions \cite{FitzHugh1955, FitzHugh1961, Nagumo1962}. 
In order to build a predictor for chaotic systems, most common techniques can be divided into the following groups \cite{Weigend}: \textit{(i)}  linear and nonlinear regression models such as  Autoregressive-Moving-Average, Multi Adaptive Regression Spline \cite{Zarandi}, and Support Vector Machine \cite{Celikyilmaz}; \textit{(ii)} state space based techniques for prediction of continuous-time chaotic systems, which utilize attractor reconstruction and interactions between internal degrees of freedom to infer the future \cite{ Kantz, Weigend, Farmer, Alparslan, Kugiumtzis}. Attractor reconstruction method is based on the embedding of the original state space in a delay-coordinate space \cite{ottcontrol}. And \textit{(iii)} the connectionist approach, including recurrent and feedforward \cite{rojas,gurney}, deep \cite{LeCunDeep}, and convolutional ANNs \cite{lecunConv}. This approach usually comprehend the design of ANNs using large amounts of neurons to process information \cite{Goodfellow:2016}.

The high-dimensionality of the ANNs' hidden layer is commonly translated in a computationally expensive problem when considering the optimization of such networks to solve a task. 
Elseways, in the reservoir computing approach such training efforts are reduced due to the training is done on the output layer only.
However, the more neurons in the hidden layer which are connected to the output layer, the higher the computational cost of the training step.  
Introducing a novel methodology, we develop a state space-based concept which guide the downsizing of the rRNNs' hidden layer. 
To achieve this objective, we describe rRNNs and state space-based models within the same framework. 
From where we show state space patterns revealed by spontaneous reconstructions inside the high dimensional space of our random recurrent network. 
Furthermore, we introduce a novel methodology based on the Takens embedding theorem to identify the embedding dimensions of the input system's spontaneous reconstruction, and their relevance to the system's prediction performance. 

We immediately exploit our insight and devise a new hybrid Takens-inspired ANN concept, in which a network is extended by an a-priori designed delay external memory. 
The delay term is used to virtually extend the size of the network by introducing virtual nodes which exist in the delay path.
We use this new design to first validate our interpretation, and then devise an advanced hybrid rRNN to stabilize a non-periodic neuronal model which requires $15$ times less neurons than a bench-mark rRNN. 
As this system is driven by a stochastic signal, we show how our approach can leverage properties of the underlying deterministic system even for the case of a stochastic drive.

\begin{figure}[t]
\centering
\includegraphics[scale=0.29]{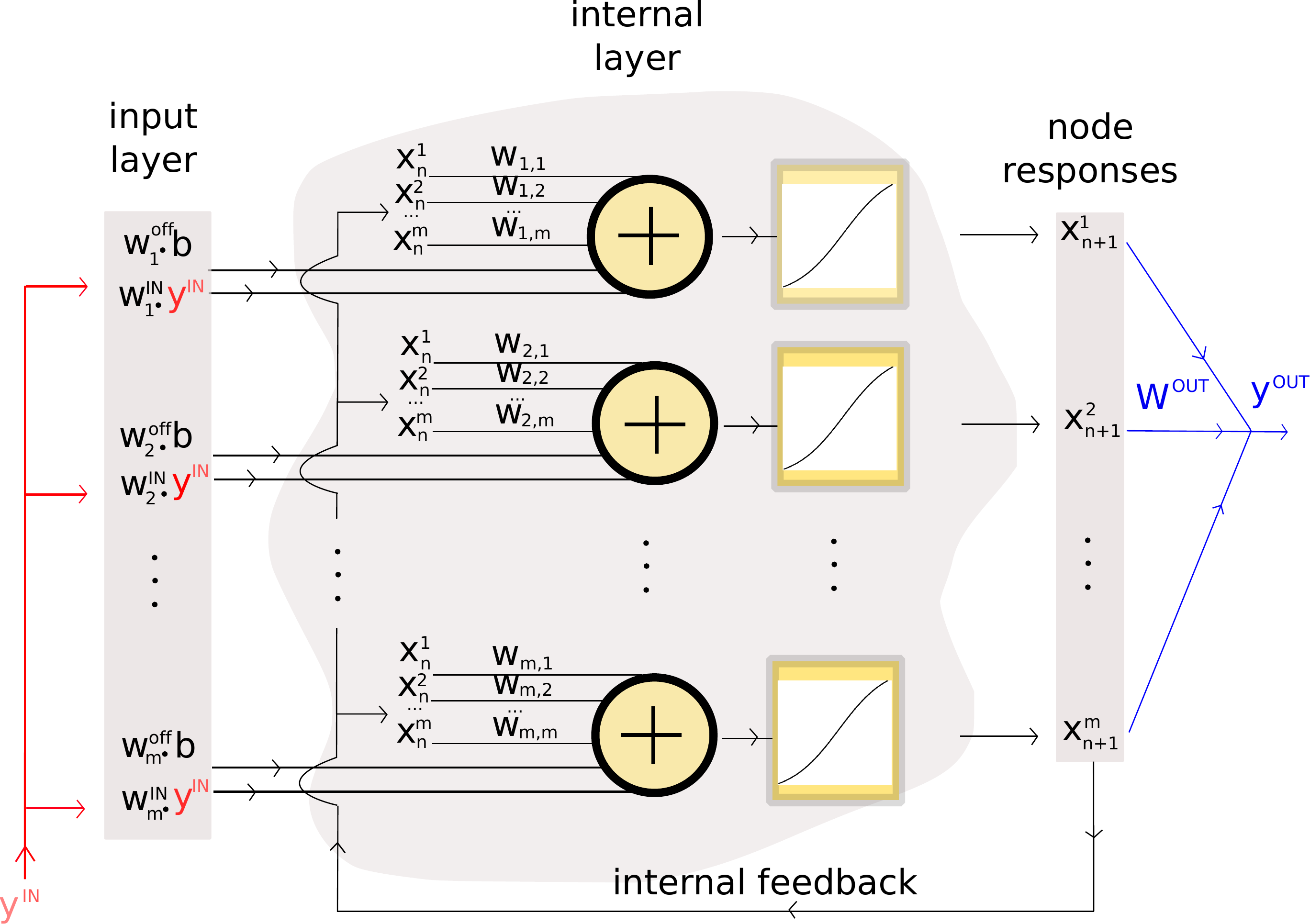}
\caption{Explicit illustration of the rRNN diagram. The network is composed by an input layer, where information $y^{IN}$ and $b$ enter to the hidden layer via random input and bias weights vectors $W^{IN}$ and $W^{off}$, respectively. The internal layer has $m$ neurons whose synaptic weights are defined by the elements of the matrix $W$. The neurons nonlinear activation functions are hyperbolic tangents. Node responses are internally fed back to the internal layer, yielding to the recurrent architecture of the network.
A readout state $y^{out}$ is created via the readout weight matrix $W^{out}$.} \label{fig:1}
\end{figure}

\section{Random Recurrent Networks for prediction}

A rRNN is illustrated in Fig.~\ref{fig:1}, indicating the temporal flow of information received by each neuron or node. Nodes are represented by $\bigoplus$. The rRNN consists of a reservoir of $m$ nodes in state $\mathbf{x}_{n}$ at integer time $n$. Nodes are connected through random, uniformly distributed internal weights defined as coefficients in the matrix $W$ of dimensionality $m\times m$. 
The resulting randomly connected network is injected with 1D input data $y^{IN}_{n+1}$ according to input weights defined as random  coefficients in the vector $W^{IN}$ of dimensionality $m\times 1$. The time-discrete equation that governs the network is \cite{Jaeger}
\begin{equation}
\mathbf{x}_{n+1}=f_{NL}(\mu W\cdot \mathbf{x}_{n} +   \alpha W^{IN}\cdot  y^{IN}_{n+1} + W^{off}\cdot b),  \label{e2}
\end{equation}
where $\mu$ is the bifurcation parameter, $\alpha$ the input gain, $f_{NL}(\cdot)$ is a nonlinear sigmoid-like activation function, and $b$ the constant phase offset injected through offset weights, defined as random coefficients in the vector $W^{off}$ of dimensionality $m\times 1$. In practice, we construct a network with $m=1000$,  using the Matlab routine \texttt{random}. Connection weights $W_{i,j}$ are distributed around zero. In order to set our recurrent network in its steady state, $W$ is normalized by its largest eigenvalue. The network's connectivity is set to one, hence it is fully connected.

An output layer creates the solution $y^{out}$ to the prediction task. In this step the network approximates the underlying deterministic law that rules the evolution of the input system. The output layer provides the computational result according to
\begin{equation}
y^{out}_{n+1}=W^{out}\cdot \mathbf{x}_{n+1}. \label{eout}
\end{equation}   
The output weights vector $W^{out}$ is calculated according to a supervised learning rule, using a reference teacher/target signal $y^{T}_{n+1}$ \cite{JaegerTechRep}. We calculate the optimal output weights vector $W^{out}_{op}$ by
\begin{equation}
W^{out}_{op}=\min_{W^{out}}\| W^{out} \cdot \mathbf{x}_{n+1} -y^{T}_{n+1}\|, \label{e2}
\end{equation}
via its pseudo-inverse (using singular value decomposition) with the Matlab routine \texttt{pinv}.
Equation (\ref{e2}) therefore minimizes the error between output $W^{out} \cdot \mathbf{x}_{n+1}$ and teacher $y^{T}_{n+1}$. As training error measure we use the normalized mean squared error (NMSE) between output $y^{out}_{n+1}$ and target signal $y^{T}_{n+1}$, normalized by the standard deviation of teacher signal $y^{T}_{n+1}$. 

When we use a rRNN for time series prediction of a chaotic oscillator such as the Mackey-Glass (MG) time-delayed system \cite{Mackey}. 
We can achieve good long-term prediction performances using a network of 1000 neurons. Here, long-term predictions are defined as predictions far beyond one step in the future. The task is to predict future steps of the chaotic MG system in its discrete-time version:
\begin{equation}
y_{n+1}=y{_n}+\delta\left(\dfrac{\vartheta y_{\tau_{m}}}{1+(y_{\tau_{m}})^{\nu}}-\psi y_{n}\right), \label{emg8}
\end{equation}
where $y_{\tau_{m}} = y_{(n-\tau_{m}/\delta)}$, $\tau_{m}=17$ as the time delay, and $\delta=1/10$ is the stepsize indicating that the time series is subsampled by 10. The other parameters are set to $\vartheta= 0.2, \nu=10, \psi=0.1$. 
For any prediction task in this paper, we consider 20-network model via different initializations of $\{W, W^{IN}, W^{off}\}$. 
The prediction horizon is estimated to be 300 time step, defined by the inverse of the largest Lyapunov exponent of the MG system ($\lambda_{max}^{MG} = 0.0036$). For such prediction horizon, and for predicting the value of 20 different MG sequences, we obtain the average of all NMSEs, resulting in $0.091\pm 0.013$. 
This performance was obtained for $b=0.2$, $\alpha = 0.8$ and $\mu = 1.1$, which was found to offer the best prediction performance in a range of $\mu\in [0.1, 1.3]$.
Moreover, the network was trained with $3000$ values of the MG system, with a teacher signal  given by $y^{T}_{n+1} = y^{IN}_{n+1}$. We subtracted the average of the MG time series before injection into the rRNN, which is a common practice \cite{JaegerTechRep}. 
Then we discarded the first $1000$ points of its response to avoid initial transients. 
Right after training, where we determined $W_{out}$, we connected $y^{out}_{n+1}$ to $y^{IN}_{n+1}$, and we left the network running freely 300 time steps, indicated by the prediction horizon.

Given that good long-term prediction performances are obtained, we wonder how input information is represented in the network's space. Such qualitative investigations are for instance common practices in image classification tasks, where the extracted features are often identified and illustrated together with the hidden layers that have generated them \cite{Taigman}.
In the following, we introduce a technique that allow us to identify feature representations of the input information in the rRNN's high dimensional space, which are linked to good prediction performances.

\section{A method for feature extraction in random recurrent networks}

As shown previously, our rRNN is able to predict the future values of 1D input chaotic data $y^{IN}$. 
Such kind of data comes from a continuous-time chaotic system. As it is known in chaos theory, a minimum of three dimensions are required in a continuous-time nonlinear system to generate chaotic solutions. 
Typically, continuous-time chaotic solutions come from models consisting of a system of at least three nonlinear ordinary differential equations (ODEs). 
However, there are other ways to obtain such chaotic dynamics. 
One of those ways includes the introduction of an explicit temporal variable to an ODE, such as a time delay with respect to the main temporal variable. We define these models as delay differential equations (DDEs), where a DDE is in fact equivalent to an infinite-dimensional system of differential equations \cite{Demidenko}.
The number of solutions to a DDE is in theory infinite due to the infinite amount of initial conditions in the continuous rank required to solve the equation. Each initial condition initializes an ODE from the infinite-dimensional system of equations. Thus, the introduction of a time delay in an ODE, resulting in a DDE, provides sufficient dimensionality to allow for the existence of chaotic solutions.
For instance, this is how the MG system can develop chaotic solutions. 
In such case, one just have access to a time series from a single accessible variable $y_{n+1}$ in Eq.~(\ref{emg8}), while all others remain hidden.
Nevertheless, hidden variables are participating in the development of the global dynamics as well as the accessible variables.

In order to approximate a full dimensional representation of these oscillators, we could embed the 1D sequence $y_{n+1}$ into a high dimensional space.
Consequently, reconstructing its state-space.
Among the most practiced methods to embed 1D information, we highlight state space reconstruction techniques such as delay reconstruction (Whitney and Takens embedding theorems \cite{Whitney, Takens}) or through the Hilbert transform \cite{Pikovsky}.

\begin{figure}[t]
\centering
\includegraphics[scale=0.8]{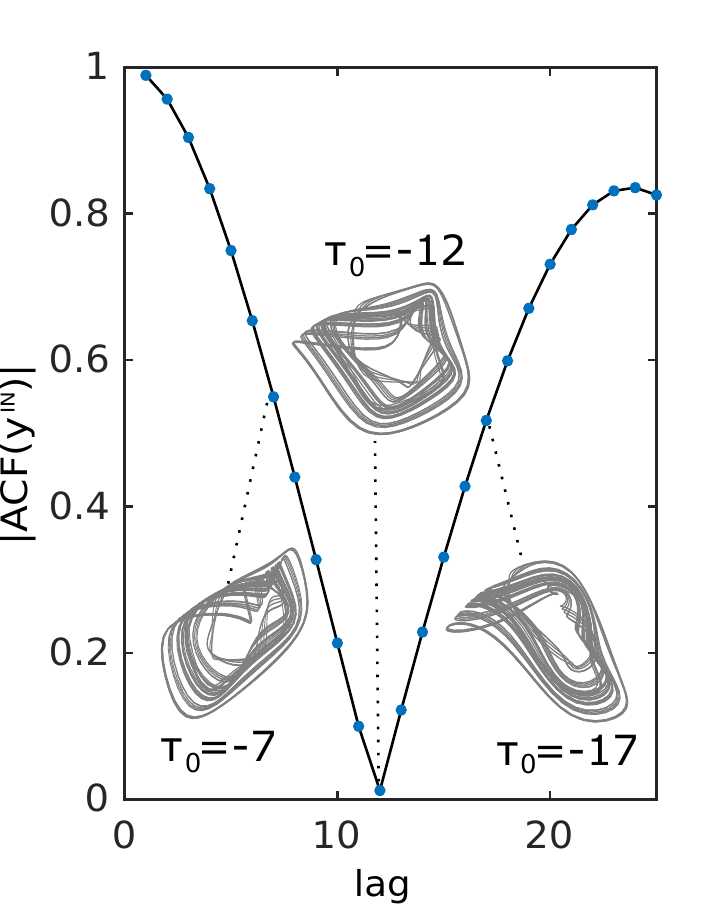}
\caption{Absolute value of the autocorrelation function for the MG system together with three examples of 2D delay-reconstruction for embedding lags $\{-7,-12,-17\}$.} \label{fig:2}
\end{figure}

Delay reconstruction is a widely used method to complete missing state space information. According to the Takens embedding theorem, the time delayed version of a time series suffices to reveal the structure of the state space trajectory. Let us represent the data in a $M$-dimensional space by the vectors  $\mathbf{y}_n=[y_{n},\; y_{(n+\tau_{0})},\; \ldots, \; y_{(n+(M-1)\tau_{0})}]^{\dagger}$, where $[\cdot]^{\dagger}$ is a transpose matrix and $y_n$ is the original time series. 
The essential parameters for state-space reconstruction are $M$ and time delay $\tau_{0}$.
Embedding delay $\tau_{0}$ is often estimated by applying autocorrelation analysis or time delayed mutual information to the signal $y_{n}$. The temporal position of the first zero \cite{Kantz} of either method maximizes the possibility to extract additional information which contains independent observations from the original signal, and hence obtain trajectories or dynamic motion along potentially orthogonal state-space dimensions. 
If successful, information of such maximized linear independence can enable the inference of the missing degrees of freedom \cite{Ruelle, Packard, Takens}. On the other hand, we estimate the minimum amount of required embedding dimensions $M$ by using the method of \textit{false nearest neighbors} \cite{Kantz}. Crucially, the following concepts are not restricted to a particular method of determining $\tau_0$ or $M$.

The autocorrelation function (ACF) is often employed to identify the temporal position $\tau_0$ used to reconstruct missing coordinates that define any continuous-time dynamical system. 
In order to show how Takens-based attractor reconstruction performs, we use a dataset of $1\times 10^{4}$ values provided by Eq.~(\ref{emg8}).
The outcome of the autocorrelation analysis reveals that the ACF has its first minimum at lag $\tau_{0}=- 12$. In Fig.~\ref{fig:2} we show the absolute value of the ACF together with three examples of  2D delay-reconstructions for three different values of the embedding lag  $\tau_0$. 
As it can be seen, $2$D-reconstructions based on lags $\tau_{0}=- 7$ and $-17$ also unfold the geometrical object within a  state space. However, aiming at a maximally orthogonal embedding, we base our analysis on the attractors reconstructed by exactly the first minimum of the ACF, $\tau_{0}=- 12$.
According to the false nearest neighbor analysis, the \textit{minimum} dimensions $M$ required to reconstruct the MG attractor is $4$. The Takens scheme therefore provides a set of coordinates  $\mathbf{y}_n = \{y_n, y_{n-12}, y_{n-24}, y_{n-36}\}$ which reconstructs the state space object.

As it can be seen, this is a classic method to extract feature representations $\mathbf{y}_n$ from the original and unknown feature $y_n$. Next, we want to identify comparable attractor reconstruction inside our rRNN's state space.

\begin{figure}[t]
\centering
\includegraphics[scale=0.8]{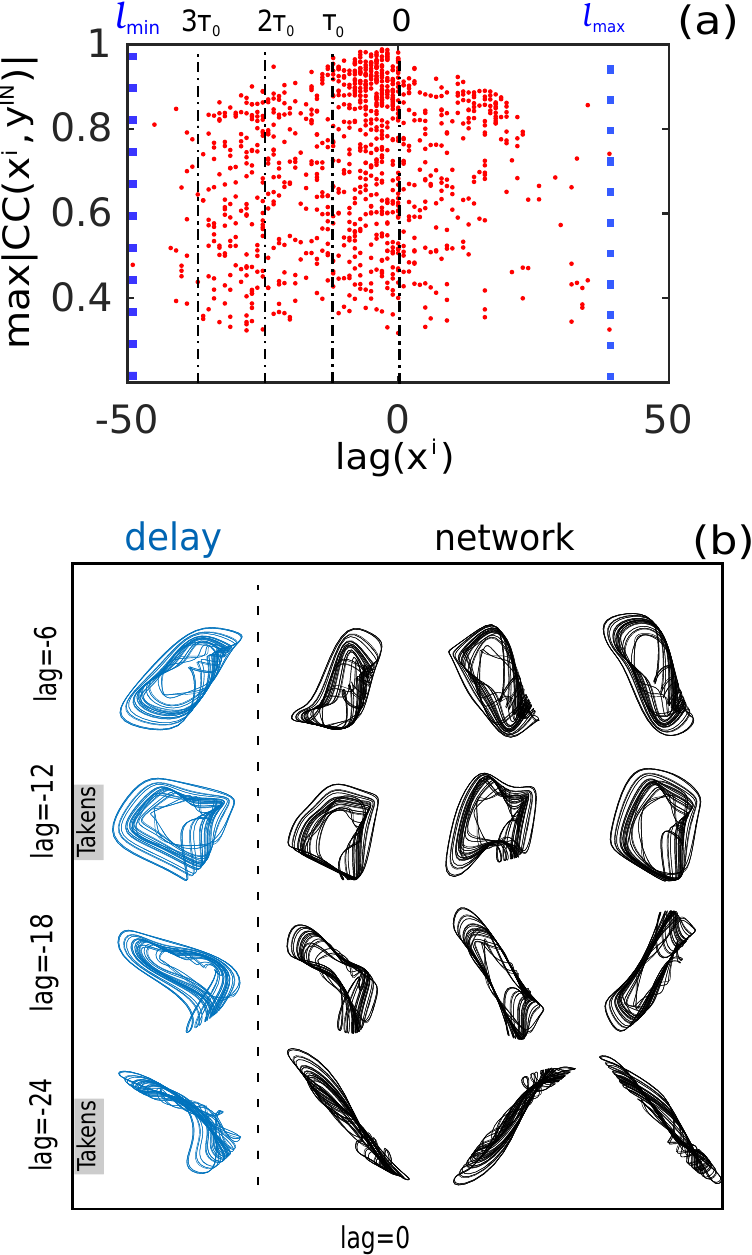}
\caption{(a) Maximum absolute value of the cross-correlation function between node responses $x^i$ and input signal $y^{IN}$ for $\mu = 1.1$, in which each of the $1000$ reservoir nodes is considered with a red dot. Labels $\{l_{min},l_{max}\}$ remark the lags' limits for the distributions of nodes, and $\tau_{0}=-12$. (b) The first column shows the 2D projection of different high-dimensional attractors of the MG system, using diverse time lags. The following three columns show the embedding found inside the rRNN's space for $\mu = 1.1$.} \label{fig:3}
\end{figure}

\begin{figure*}[t]
\centering
\includegraphics[scale=0.71]{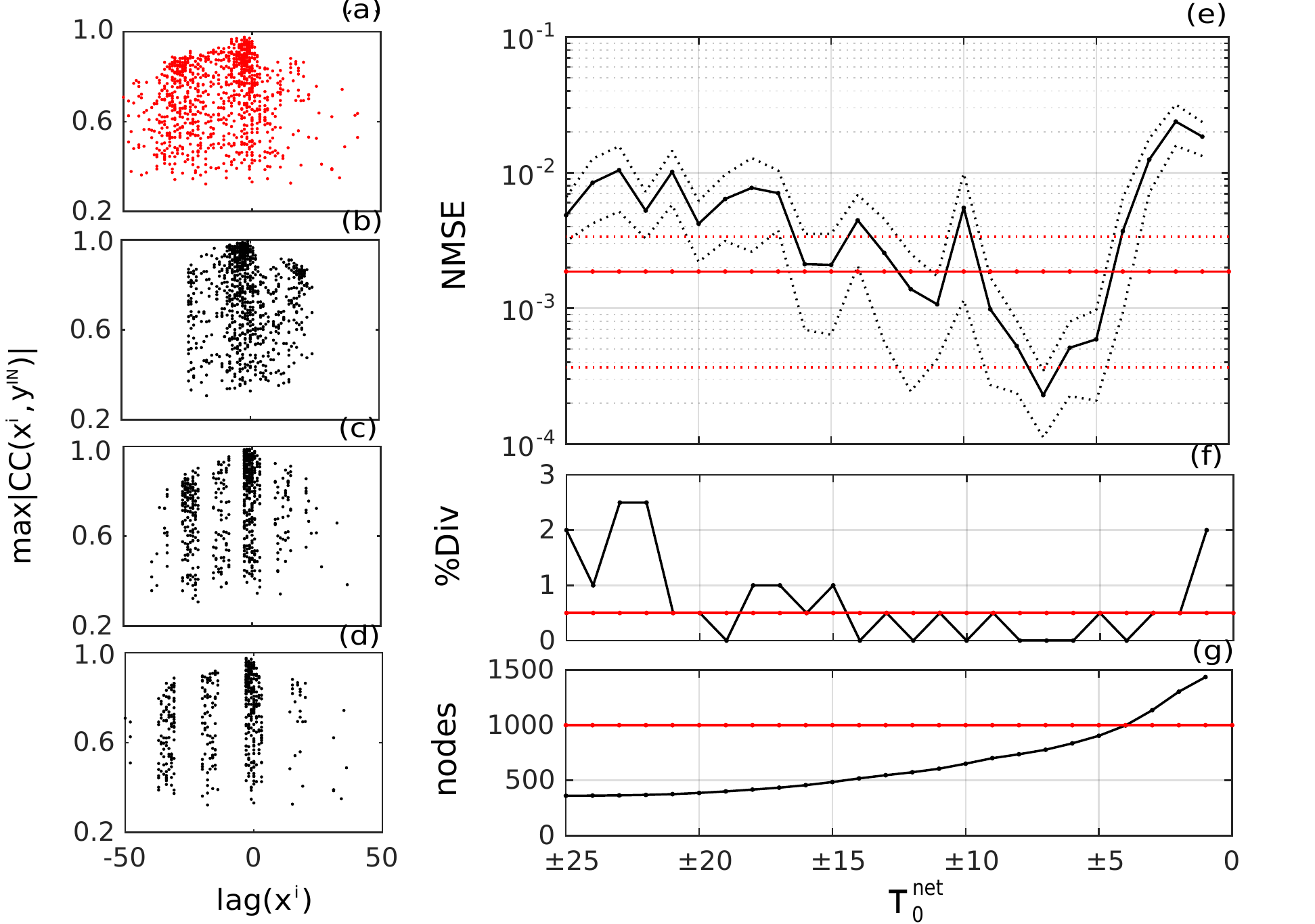}
\caption{Maximum absolute value of the cross-correlation function between node responses and input signal for $\mu = 1.1$, considering the rRNN network with: (a) all nodes; and nodes lagged at $\Gamma(\tau_{0}^{net})=\{(M-1)\tau_{0}^{net}\pm\delta\tau_{0}^{net}\}_{M}$, where $M=1,2,3,4$ and $\delta\tau_{0}^{net}=3$, for (b) $\tau_{0}^{net}=- 7$, (c) $\tau_{0}^{net}=- 12$ and (d) $\tau_{0}^{net}=- 17$. (e) Measure of prediction error via average of NMSEs for $20$-network model at $\mu = 1.1$, over $20$ MG different time series. The red constant lines shows prediction performances for the networks including all nodes. The black curves shows prediction performances only considering nodes contained in $\Gamma(\tau_{0}^{net})$. (f) Percentage of divergence showing the average of realizations for which NMSE$>1$, and (g) average of nodes contained in $\Gamma(\tau_0^{net})$. } \label{fig:4}
\end{figure*} 

\subsection{A Takens-inspired feature extraction technique}

The network's state space is defined by the set of orthogonal vectors in $W$.
For that, we analyze the injected signal representation $y^{IN}_{n+1}$ via network's node responses.
As input information $y^{IN}_{n+1}$ is being randomly injected into the network's high dimensional space, we search for possible spatial representations of the 1D input, where network nodes serve as embedding dimensions. 

With that goal in mind, we proceed with an analysis comparable to the ACF used in delay embedding, based on the maximum absolute value of the cross-correlation function $\vert CC(x^{i},y^{IN}) \vert_{max}$ between all node responses $\{x^{i}\}$ and the input data $y^{IN}$. 
As our aim is to identify nodes providing observations approximately orthogonal to $y^{IN}$, we record for every network node such cross-correlation maxima and the temporal position, or lag, of this maxima. Figure~\ref{fig:3}(a) shows the cross-correlation analysis (CCAs) for $\mu=1.1$, where node correlation lags and maxima correspond to the abscissa and ordinate, respectively. 
In this case, a distribution of node lags, where an extension of the range to $\{l_{min},l_{max}\}=\{-49,45\}$ is shown. Thus, the rRNN's nodes reveal a strong cross-correlation at time lags covering all Takens embedding delays $\{0, \tau_{0},2\tau_{0},$ $3\tau_{0} \}$. Nodes lagged around $\{- 36, - 24,- 12,0\}$ should well approximate the Takens embedded attractor, and a state space representation of the input sequence is presented within the rRNNs' space.  

In Fig.~\ref{fig:3}(b), we illustrate the numerous possible extracted features from the originally injected attractor embedded in the rRNN's space for $\mu=1.1$. The first column of the figure shows simple $2$D-projections of the delay-reconstructed attractor by using the set of lags $\{ -24,-18,-12,-6 \}$. 
The attractors reconstructed by network nodes lagged at $\{ -24,-18,-12,-6 \}$ are shown in the three next columns. 
Attractors reconstructed by network nodes with maximum cross-correlation values at lags $\{ -18,-6 \}$ do not belong to the set of original Takens delay-coordinates. We included these additional delay-dimensions to better illustrate the breadth of features provided by the rRNN.

As shown above, the CCA also finds node lags and correlations that do not agree with the Takens framework. 
In the following, we introduce a methodology to exclude such additional delay-dimensions from our predictor.
As an starting point, we suppress nodes with specific CCA-lag positions during the training step of the output layer, such that they will not be available to the readout matrix but still take part in the rRNN's state evolution.
To that end, we select nodes for which their CCA-lag positions are within windows of width $\delta\tau_{0}^{net}$, centered at integer multiples of $\tau_{0}^{net}$, where $\tau_{0}^{net}$ represents the time lag used for delay reconstructions in the Takens' scheme. 
The windows width $\delta\tau_{0}^{net}$ defines a time-lag uncertainty associated to the identified rRNNs' delay coordinates.
All nodes with CCA-lag positions not inside the set of ($n\tau_{0}^{net}\pm\delta\tau_{0}^{net})$, $n\in\mathbb{Z}$ will not be available to the readout layer.

To illustrate our method and its effect, we show the non-filtered CCA of the rRNN when driven by the MG signal and with bifurcation parameter $\mu = 1.1$  in Fig.~\ref{fig:4}(a). Using a constant CCA-windows width of $\delta\tau_0^{net}= 3$, as the minimum uncertainty found associated with good performances, we now scan the position of these CCA-windows by changing $\tau_0^{net}$. We restrict the number of windows to $n\in\{-M, -M+1, \dots, M\}$. For $\tau_0^{net}\in\{-17, -12, -7\}$, in panels (b-d) of Fig.~\ref{fig:4}, we show examples of filtered CCAs where just rRNN nodes available for the readout layer are present.

Based on such movable CCA-filters, we can estimate the relevance of different CCA lags on the rRNN's capacity to predict a particular temporal sequence. We define 20-network model via different initializations of $\{W, W^{IN}, W^{off}\}$, and for each we obtain the NMSE for predicting the value of 20 different MG sequences at 300 time steps into the future. In Fig.~\ref{fig:4}(e), we show the resulting NMSE, averaging over all system combinations and for $-25 \leq \tau_0^{net} \leq -1$. Here, the average NMSE is given by the solid line, and the standard deviation (stdev) interval by the dashed black lines. In panel (f) we show the percentage of rRNN's for which prediction diverged from the target, i.e. NMSE$>1$. Panel (g) shows how many nodes are available to the network's output layer.
The constant red curves present in all panels show averaged performances obtained for the non-filtered rRNN, namely the classical RC, again with the stdev interval given by the dashed red lines.
 
Restricting the system's output based on the CCA-filter windows has a strong and systematic impact onto the rRNN's prediction performance. Performance is optimized for very characteristic filter positions, i.e. for $\tau_0^{net}\in[-16,-5]$. For $\tau_0^{net}\in [-12,-11]$ the embedding available to the network's output closely corresponds with the lags of the original Takens attractor embedding. 
The performance achieved by setting $\tau_{0}^{net} \simeq \tau_{0}$ in our approach reduces the NMSE more than 1-fold, even though the system has in average significantly less network nodes available to the output layer ($\sim500$), see Fig.~\ref{fig:4}(g). 
For $\tau_{0}^{net} = -7$, the performance is higher by one order of magnitude, accessed by using approximately the same initial set of nodes available to the output ($\sim1000$), see Fig.~\ref{fig:4}(g).
We are therefore able to identify node families that shows attractor embedding features in the network's space for the first time, based on their CCA-lag. 

In summary, it can be seen that for filters based on $\tau_0^{net}\in[-16,-5]$ the prediction performances are either coincident or better than the non-filtered CCA case. This result is in agreement with Fig.~\ref{fig:2}, where we show that lags in such interval seem to still unfold the object in the state space.

Finally, some further aspects are seen in our data.
For $\tau_0^{net}> -4$ the CCA-filter windows overlap and nodes with a lag inside such positions of overlap are assigned to multiple windows. This artificially increases the number of nodes available to the system's output beyond $m=1000$. The resulting NMSE strongly increases beyond the one for the original rRNN. We attribute this characteristic to over-fitting during learning, where there are just repetitions of the same few delay-coordinates.

\subsection{Characteristics of the extracted features}

  \begin{figure*}[t]
\centering
\includegraphics[scale=0.67]{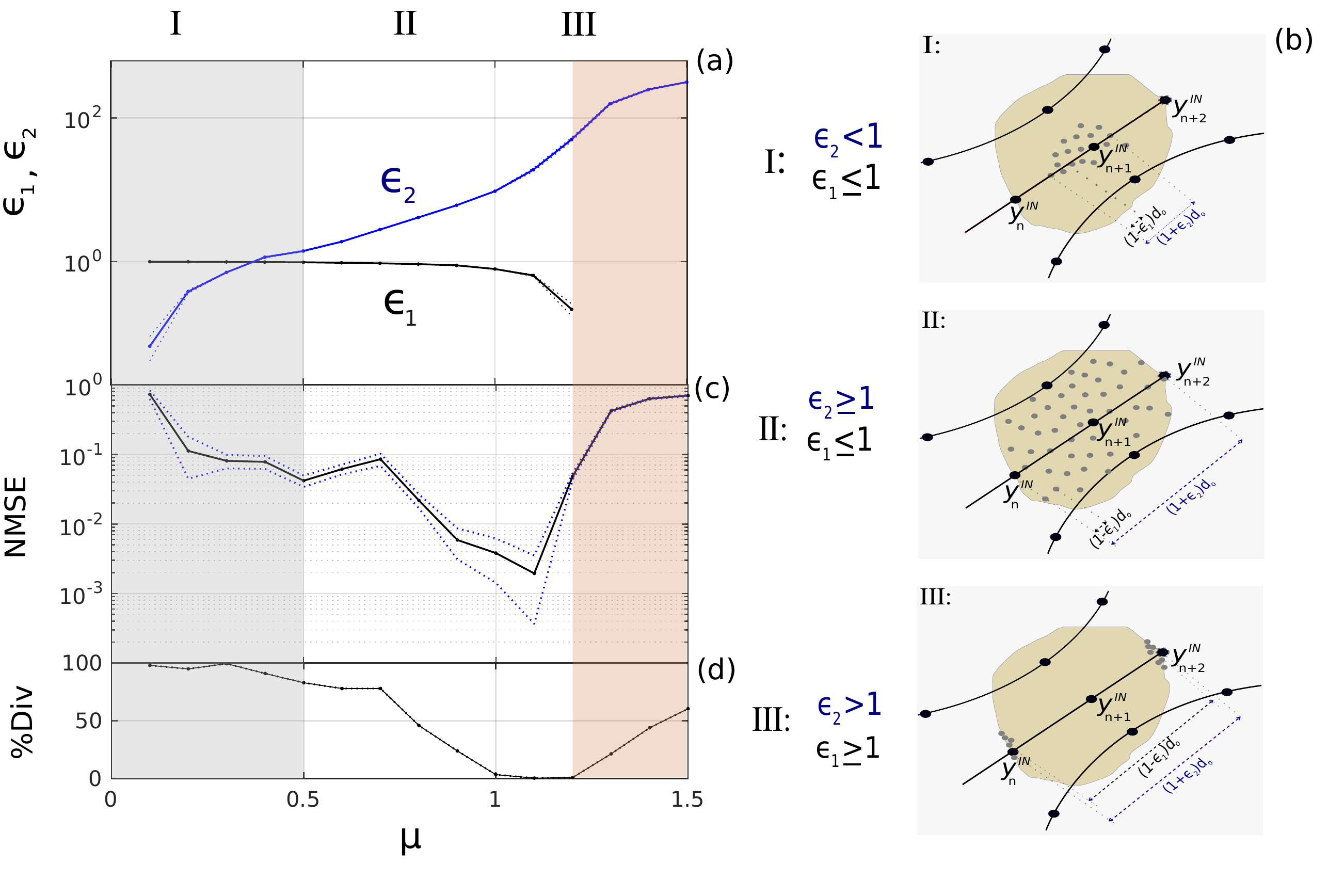}
\caption{ (a) Maximum and minimum average boundaries $\{ \epsilon_{1}, \epsilon_{2} \}$ as function of $\mu$, where  $\tau_0^{net}=-12$ and $\delta\tau_0^{net}=3$. (b) Illustrative scheme showing the evolution of $\{ \epsilon_{1}, \epsilon_{2} \}$ for three different cases connected to their corresponding ranges in $\mu$. (c) Measure of prediction error via average of NMSEs for $20$-network model, over $20$ MG different time series; and (d) percentage of divergence showing the average of realizations for which NMSE$>1$ as functions of $\mu$. } \label{fig:44}
\end{figure*}

 \begin{figure}[t]
\centering
\includegraphics[scale=0.75]{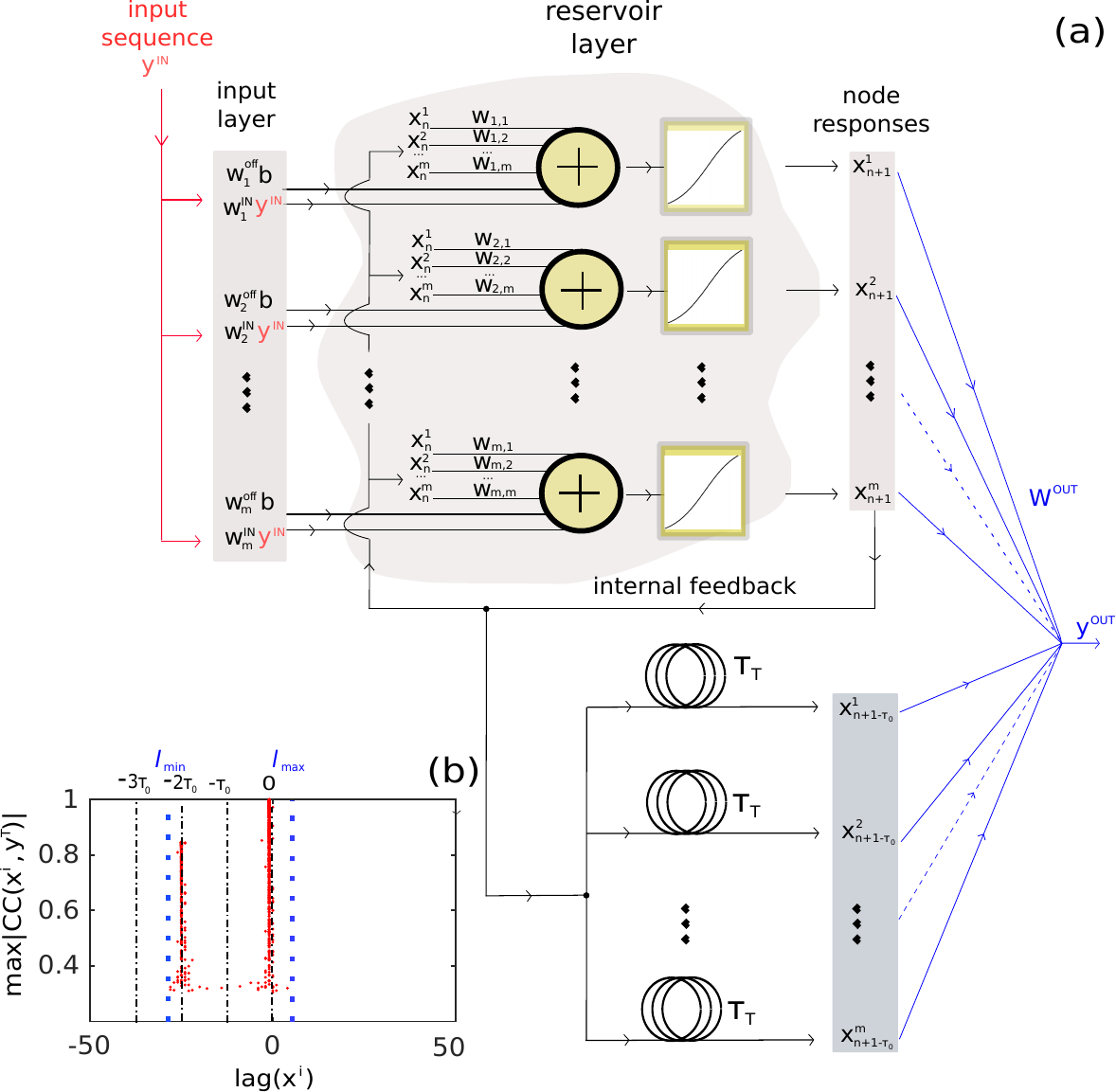}
\caption{(a) Schematic illustration of an TrRNN. Information enters the system via the input, a recurrently connected network forms a neural network. Based on our theory, we propose a simplistic extension to the system via an external delay memory $\tau_{T}$. (b) Maximum absolute value of the cross-correlation function between node responses $x^i$ and input signal $y^{IN}$ for $\mu = 0.1$, in which each of the $1000$ reservoir nodes is considered with a red dot. Labels $\{l_{min},l_{max}\}$ remark the lags' limits for the distributions of nodes, and $\tau_{0}=-12$.} \label{fig:t}
\end{figure} 

\begin{figure}[t]
\centering
\includegraphics[scale=0.6]{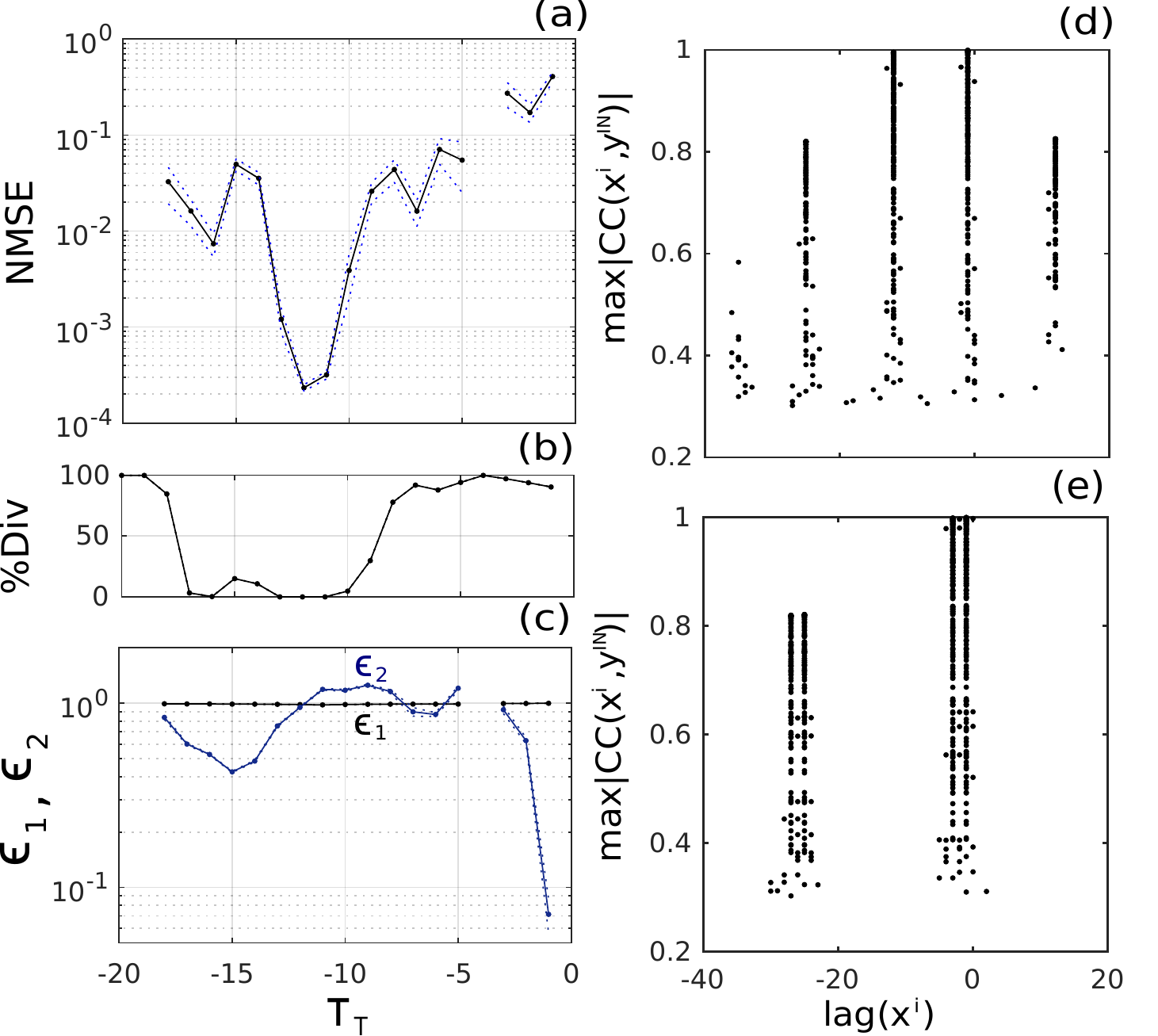}
\caption{(a) Average of prediction performances NMSEs for $20$-network model, over $20$ MG different time series using a TrRNN with $350$ nodes at $\mu =0.1$ and the delay term $\tau_{T}\in [-20,0]$. (b) Percentage of divergence showing the average of realizations for which NMSE$>1$. (c) Maximum and minimum average boundaries $\{ \epsilon_{1}, \epsilon_{2} \}$ as function of $\tau_{T}$. Maximum absolute value of the cross-correlation function between node responses and input signal for (d) $\tau_{T}=-12$ and (e) $\tau_{T}=-3$. } \label{fig:5}
\end{figure}

Our previous analysis identifies and harness meaningful features related to  good prediction performances. As this was realized by randomly connected networks, attractor reconstruction was achieved by randomly mapping the originally 1D-input onto the high-dimensional rRNN's space.
Such random mapping is treated by the framework of random projections theory (RPT) \cite{JLlema, Indyk, Frankl, Dasgupta, Sivakumar}.
An active area of study within this field treats the question if original input data is randomly mapped onto the dimensions of the projection space, the structural damage to the original object is minimized.
 
To determine the degree of potential structural distortions to the original input attractor after random mapping onto the network's high-dimensional space according to $\mathbf{y}^{IN}_n \rightarrow \varphi(\mathbf{y}^{IN}_n)$, we measure distances between consecutive states (interstate distances) of the original $||\mathbf{y}_{n+1}^{IN} - \mathbf{y}_{n}^{IN}||$ and the projected $||\varphi(\mathbf{y}_{n+1}^{IN}) - \varphi(\mathbf{y}_{n}^{IN})||$ objects. 
 Following these steps, we take inspiration from RPT extended to nonlinear mapping \cite{Tenenbaum}. 
Under such mapping, the interstate distances of the original attractor $\| \mathbf{y}^{IN}_{n+1}- \mathbf{y}^{IN}_{n} \|$, and of the projected attractors $\| \varphi(\mathbf{y}^{IN}_{n+1})- \varphi(\mathbf{y}^{IN}_{n}) \|$ in the rRNN's space, are bound to the range $[(1-\epsilon_{1}),(1+\epsilon_{2})]$ according to:
\begin{widetext}
\begin{equation}
(1-\epsilon_{1})\| \mathbf{y}^{IN}_{n+1}- \mathbf{y}^{IN}_{n} \| \leq \| \varphi(\mathbf{y}^{IN}_{n+1})- \varphi(\mathbf{y}^{IN}_{n}) \| \leq (1+\epsilon_{2})\| \mathbf{y}^{IN}_{n+1}- \mathbf{y}^{IN}_{n} \|. \label{JLb1}
\end{equation} 
\end{widetext}
where $\{\varphi(\mathbf{y}^{IN}_{n}), \varphi(\mathbf{y}^{IN}_{n+1})\}$ are states built by rRNN node responses, and assigned to an embedding dimension using the CCA.
Details in the estimation of $\{ \epsilon_{1}, \epsilon_{2} \}$ are added in Appendix A.

As $\tau_{0}^{net}=-12$ was found to be associated to a good prediction performance using approximately half of the initial set of nodes, we show in Fig.~\ref{fig:44}(a) the estimation of $\{ \epsilon_{1}(\mu), \epsilon_{2}(\mu) \}$ for the rRNN for which nodes' CCA-lag positions are within windows of width $\delta\tau_{0}^{net}=3$, centered at integer multiples of $\tau_{0}^{net}=-12$. 
Here, we present the average of the statistical distribution that includes $20$-network models and predicting 20 different MG time series at 300 time steps into the future. In panel (b), we schematically illustrate the relevant geometrical properties of the attractors mapped onto the rRNN's space.
Such study considered $\tau_{0}=-12$ with which we obtain the set of coordinates  $\mathbf{y}_n = \{y_n, y_{n-12}, y_{n-24}, y_{n-36}\}$ that we use to unfold the state space object; where $\mathbf{y}^{IN}_n = \mathbf{y}_n$. 

The consequence of an increasing $\mu$ in Fig.~\ref{fig:44}(a) can be explained with the graphic representation of the limits shown by Fig.~\ref{fig:44}(b). Here, we illustrate the three general cases (I, II, III) connected to their corresponding ranges in $\mu$ in Fig.~\ref{fig:44}(a). 
The evolution of the system's trajectory is illustrated along a curve sampled at the positions of the big black dots. Gray dots are network's neighbor states to the state $y^{IN}_{n+1}$.
The first case (I) corresponds to neighbors which form a dense cloud of samples since $\epsilon_1\sim 1$, but are arranged closely around the original sample since $\epsilon_{1} \lesssim 1$ and $\epsilon_{2} <1$. 
The neighbor samples insufficiently enhance diversity in feature extraction and the network cannot predict the system's future evolution. This is confirmed by Fig.~\ref{fig:44}(c,d), which show bad prediction performance and unity divergence: the RNN cannot predict the system.

Case (II) includes the values of $\mu$ where $\epsilon_{1} \simeq 1$ and $\epsilon_{2} \geq 1$. 
Within this parameter range, the system's prediction performance strongly increases until reaching the lowest prediction error. Our analysis reveals the following mechanism behind this improvement. According to $\epsilon_2$,  the maximum interstate distance possible inside the rRNN's space is twice the interstate distance of the original trajectory. As a consequence, the rRNN samples  neighbors to state $\mathbf{y}^{IN}_{n+1}$. Hence, as the state neighborhood is broader, there now is a sufficient random scanning of the attractor's vicinity, such that the network can use the different features to solve prediction. The network can therefore use the projected objects to predict, which is confirmed by excellent performance according to Fig.~\ref{fig:44}(c,d).

The last case (III) appears for $\mu>1.3$, where all approximated distances of the embedded attractor are much larger than the original distance.
The rRNN's autonomous dynamics therefore enlarge the sampling distance such that no dense nearest neighbors are anymore available for prediction.
The result are the distortions caused by the autonomous network's dynamics typically found to be chaotic for $\mu \gtrsim 1.4$, already identified in our previous work \cite{Marquez2018}. 
Consequently, the network's folding property distorts the projected features. Therefore, $\epsilon_{1}$ becomes undefined, meaning that information about the structure of the embedded trajectory is lost. As a consequence, prediction performance strongly reduces, see Fig.~\ref{fig:44}(c,d). 

The process described in this section allowed us to identify and use relevant features that benefit good long-term prediction performances with a downsized rRNN. Additionally, we found that the neighborhood generated by the inter-state distances of such features has an meaningful impact on the network's ability to predict at all. 
At this point, we show how to simplify even more our process and package all the above described steps that allow us to generate features related to good prediction performances.

 \begin{figure*}[t]
\centering
\includegraphics[scale=0.7]{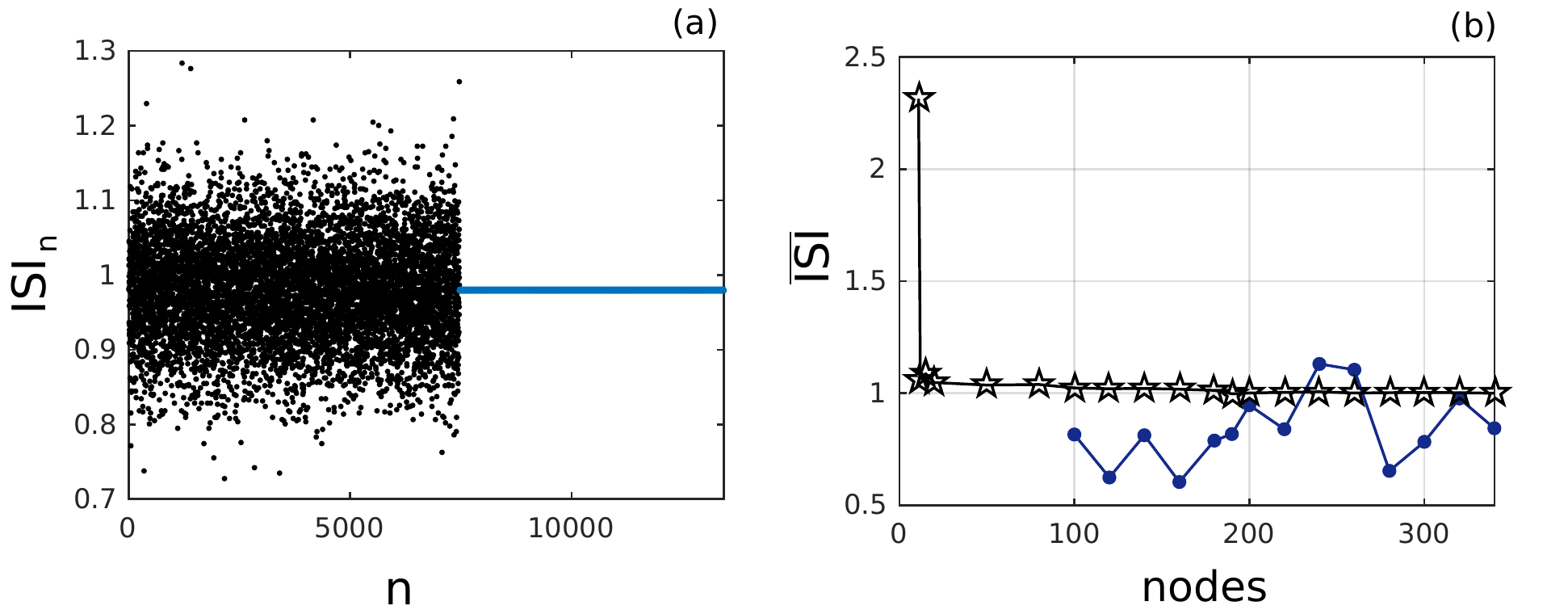}
\caption{(a) Interspike intervals (ISI) of a arrhythmic excitable system comparable to a heart.
Stabilization of the system based on TrRNN with only $12$ network nodes. (b) Comparison between the stabilized mean of the TrRNN (black curve with stars) and a classical rRNN (blue curve with dots). } \label{fig:6}
\end{figure*}

\section{Hybrid Takens-inspired random recurrent networks }

In this section, we directly exploit our newly gained insight and introduce a modified version of the classical random neural network for time series prediction.
Here, we design a system which aims to only take into account such Takens-like dimensions that we found to be relevant for prediction. 
As it was previously described, actions provided by the nodes of the rRNN can be interpreted in the light of delay-embedding.
We consequently modify the classical rRNN by including a Takens-inspired external memory:
\begin{eqnarray}
\textbf{x}_{n+1}=f_{NL}(\mu W\cdot \textbf{x}_{n} +  \alpha W^{IN}\cdot y^{IN}_{n+1} + W^{off}\cdot b), \\
y^{out}_{n+1} =W^{out} \cdot (\textbf{x}_{n+1},\textbf{x}_{n+1+\tau_{T}}) \label{e2a}
\end{eqnarray}
where $\textbf{x}_{n+1+\tau_{T}}$ is a delayed term added to the output layer, see Fig.~\ref{fig:t}(a). All elements of the reservoir layer have been  copied and then time shifted by a delay term $\tau_{T}$ that could be the Takens embedding delay  $\tau_{0}$. This process allows us to add virtual nodes to our network, which are distributed in the delay lines. This Takens rRNN (TrRNN) combines nonvolatile external memory (virtual nodes) with a neural network (real nodes) and therefore shares functional features of the recently introduced hybrid computing concept \cite{Graves2016}. Yet, our concept makes any additional costly optimization unnecessary. 

We start our analysis by identifying the embedding delay related to the best prediction performance.
Here, we fix $\mu =0.1$, and modify then the delay term $\tau_{T}\in [-20,-1]$.
For $\mu =0.1$, the delay-coordinates found in the network's space only span approximately two Takens-like embedding dimensions of the MG system with delays $\{2\tau_{0},0 \}$, when $\tau_{0}=-12$, see Fig.~\ref{fig:t}(b).  
Furthermore, most node responses are distributed along the columns centered in lags $\{2\tau_{0},0 \}$. 
Consequently, as shown in Fig.~\ref{fig:44}(c,d), the prediction performance is almost the lowest possible due to insufficient dimensionality to get attractor-like features. 
Additionally, we set the number of nodes to $350$, which is $70\%$ of the nodes that the best CCA-windows filtered rRNN had to disposal, see Sec.~III.A; and it uses $35\%$ of the nodes used by the non-filtered classical rRNN. 

According to Fig.~\ref{fig:5}(a,b), the best average prediction performance, for 20-network model over 20 MG different time series at 300 time steps into the future, is found for $\tau_{T}=-12$, belonging to an interval $\tau_{T}\in[-13,-10]$  with the lowest NMSE values and divergent rates. 
The delay $\tau_{T}$ therefore agrees with the one identified for Takens attractor embedding $\tau_{0}$ and is almost identical to one of the lags $\tau_{0}^{net}$ found optimal in the CCA-window filtering. Furthermore, here the system only has as many CCA-windows at its disposal as dimensions required to embed the MG attractor.
This lifts the disambiguate present in the CCA-window filtering analysis, and consequently the optimum performance is found only for an TrRNN embedding exactly along lags according to Takens embedding. 
In comparison to the classical rRNN, our TrRNN achieves the same performance, simultaneously reducing the amount of nodes in the reservoir layer from $1000$ to $350$ in the output when compared to the CCA filtered system. Compared to the pristine rRNN, we obtain one order of magnitude better performance with a network three times smaller.

Figure~\ref{fig:5}(c) shows the estimation of $\{\epsilon_{1}, \epsilon_{2}\}$ with the variation of $\tau_{T}$.  As it can be seen, $\epsilon_{2}\approx1 $ is associated to good prediction performances, found for $\tau_{T}\in[-13,-10]$. This result agrees with the results provided by the classical random network in Sec.~III.B. The CCA for $\tau_{T}=-12$ is shown by Fig.~\ref{fig:5}(d), where we can find the set of nodes with all delay-coordinates required to fully reconstruct the MG attractor. In the cases where prediction was not possible, the CCA identifies the non-adequacy of the rRNN delay embedding as the reason, see Fig.~\ref{fig:5}(e) for $\tau_{T}=-3$.

\subsection{Application: control an arrhythmic neuronal model}

We directly utilize our TrRNN as a part of an efficient feedback control mechanism in an arrhythmic excitable system.
We task the TrRNN to aid stabilizing a system which models the firing behavior of a noise-driven neuron. It consists in the FitzHugh-Nagumo (FHN) neuronal model \cite{Longtin, Christini},
\begin{eqnarray}
\epsilon \dfrac{dv(t)}{dt}= v(t)[v(t)-g][1-v(t)]-w+I+\xi(t), \\ \label{FHN}
\dfrac{dw(t)}{dt}=v(t)-Dw(t)-H,
\end{eqnarray}
where $v(t)$ and $w(t)$ are voltage  and recovery variables. $I=0.3$ is a tonic activation signal, $\xi$ is Gaussian white noise with zero mean and standard deviation $\sim 0.02$, $\epsilon=0.005$, $g=0.5$, $D=1.0$, and $H=0.15$. These equations have been solved by the Euler-Maruyama algorithm for stochastic differential equation's integration. In its resting state, the neuron's membrane potential is slightly negative. Once the membrane voltage $v(t)$ is sufficiently depolarized through an external stimuli, the neuron spikes due to the rise of the action potential \cite{Enoka, Tirozzi}. The time between consecutive spikes are defined as \textit{interspike intervals} (ISIs).  These random ISIs are shown by dots in black in Fig.~\ref{fig:6}(a), clearly showing that spiking is non-regular.

We aim to control this random spiking behavior of the FHN neuronal model by proportional perturbation feedback (PPF) method \cite{Garfinkel} and by using either a TrRNN or a rRNN. The PPF method consists in the application of perturbations to locate the system's unstable fixed point onto a stable trajectory \cite{Garfinkel, Schiff}. This method is used to fit instabilities in the FHN neuronal model through the design of a control equation. In our case, the goal of using the PPF method is to build a control subsystem, which applies an external stimuli to trigger spiking and reduce the degree of chaos.

In our approach, the past information provided by voltage $v(t)$ in the FHN model is used to determine two things: \textit{(i)} the parameters to design the control equation, and \textit{(ii)} training parameters for rRNN ($\mu=1.1$) and TrRNN ($\mu=0.1$, and $\tau_{T}=-166$ obtained via the ACF minimum of $v(t)$ as in the Takens' scheme) to predict future values of $v(t)$. The predicted $v(t)$ is used to calculate the full control signal with which we stabilize the neuron's spiking activity. This methodology allows us to replace the quantity under control $v(t)$ by a reconstructed signal, which in control theory is related with the replacement of \textit{sensors} in an exemplary control system.
 To train the network, we inject $1\times 10^{5}$ values and we let the network run freely for other $4\times 10^{6}$ steps, allowing us to stabilize $5619$ ISI points. We then evaluate the quality of the stabilization for networks ranging from $11$ to $340$ nodes. 
In Fig.~\ref{fig:6}(a), the set of blue dots along a constant line shows how the TrRNN can stabilize the ISI activity. As it can be seen, the network can control the random ISI starting from $n=7465$. The excellent stabilization was achieved with TrRNN with only $12$ nodes.

Figure~\ref{fig:6}(b) shows the full comparison between rRNN and TrRNN. The mean value of ISI is calculated for the different sizes of rRNN and TrRNN and then normalized by the mean value of the random ISI. The TrRNN starts inferring the inner dynamics of the FHN system for an extremely small network containing just $12$ nodes, from which point on it is always capable to correctly stabilize the ISI. In contrast, the classical rRNN does not predict at all until its architecture has at least $80$ nodes, but performance remains poor in comparison with TrRNN. For $200$ nodes the rRNN starts predicting the dynamic of the FHN system more or less correctly, allowing the control signal to fully stabilize the ISI. Yet, for more than $200$ nodes the good performance still can fluctuate, even significantly dropping again. This indicates that in general the stabilization via a classical rRNN is not robust. Furthermore, with the TrRNN one can reduce the number of nodes to $15$ times less than the classical rRNN. This stark difference in performance highlights \textit{(i)} the difficulty of the task, and \textit{(iI)} the excellent efficiency that the addition of a simple, linear delay term adjusted to the Takens embedding delay brings to the system. Our TrRNN therefore is not only a interesting novel ANN-concept for the prediction of complex systems. It also once more confirms our theory: embedding inside ANNs is essential if the systems are to be used for prediction.

\section{Comparison with classical state space prediction}

Up to now, we have been describing a method for downsizing the hidden layer of rRNNs which can be summarized by the same steps that should be followed if we attempt to solve continuous-time chaotic signal prediction in the state space framework. 
This framework can be divided according to three fundamental aspects \cite{Weigend, Kantz, Farmer}: \textit{(i)} insufficient information to represent the complete state space trajectory of the chaotic system. 
This problem originates from the fact that in many cases one does not have access to all state space dimensions. 
In this case a reconstruction of the dynamics along the chaotic system's missing degrees of freedom is required. The knowledge of all dimensions allows us to design predictors based on the full state space trajectories. 
\textit{(ii)} The second problem is related to the sampling resolution. All information that is acquired, be it from simulations or from experiments, comes with a particular resolution. To minimize the divergence between a prediction and the correct value, the sampling resolution has to be maximized. This is of particular importance for prediction of chaotic systems as these by definition show exponential divergence.
\textit{(iii)} For deterministic chaotic systems, future states of a given trajectory can in principle be approximated from the exact knowledge of the present state. Therefore, the final step towards prediction is approximating the underlying deterministic law ruling the dynamical system's evolution. 

Therefore, step \textit{(i)} is fulfilled by the random mapping which takes place in the high dimensional space of the network. Here, RPT supports the fact that the original input data are randomly mapped onto the dimensions of the projection space, and then the structural damage to the original object is minimized. 
Step \textit{(ii)} is fulfilled by the analysis made in Section~III.B, where the sampling resolution is maximized to cover the region between the states $\mathbf{y}^{IN}_n$ and $\mathbf{y}^{IN}_{n+1}$. 
Finally, \textit{(iii)} relates to the training itself of the rRNN, where $W^{out}$ has to be determined via regression.

\section{Conclusion}

We have introduced a new method of rRNNs analysis which demonstrates how prediction is potentially achieved in high dimensional nonlinear dynamical systems. Random recurrent networks and prediction of a specific signal can consequently be described via a common methodology. Quantifying measures such as the memory related cross-correlation analysis and the feature extraction are quantitatively interpretable. We therefore significantly extend the toolkit previously available for random neural network analysis. Tools developed in the article might be comparable to the utilization of the t-SNE \cite{Maaten} technique for analyzing ANNs during a classification task.

Our scheme has numerous practical implications. The most direct is motivating the development and analysis of new learning strategies. Furthermore, we already designed a novel hybrid-computer which includes both, virtual and real nodes, that efficiently predicts via a priori defined external memory access rules. 
This approach allows us to improve the design of our neural network in order to reduce the number of nodes and connections required to solve prediction. 
Finally, our work partially removes the black-box property of random recurrent networks for prediction, possibly giving translational insight into how such tasks can be solved in comparable systems.

\section*{Acknowledgements}
Funding for B.A.M. and B.J.S. was provided by the 2019 Queen's postdoctoral fellowship fund, the Natural Sciences and Engineering Research Council of Canada (NSERC), and the Queen’s Research Initiation Grant (RIG).  

\appendix

\section{Estimation of $\{ \epsilon_{1}, \epsilon_{2} \}$}

Each state in Takens space is described by $M$ delay-coordinates,
\begin{eqnarray}
\mathbf{y}^{IN}_{n} = (y^{IN}_{n},y^{IN}_{n+\tau_{0}}, \ldots, y^{IN}_{n+(M-1)\tau_{0}}), \\ \mathbf{y}^{IN}_{n+1} = (y^{IN}_{n+1}, y^{IN}_{(n+1)+\tau_{0}}, \ldots, y^{IN}_{(n+1)+(M-1)\tau_{0}}).  
\end{eqnarray}
The second step is to define the corresponding two arbitrary consecutive states  $\{\varphi(\mathbf{y}^{IN}_{n}),$ $ \varphi(\mathbf{y}^{IN}_{n+1}) \}\in \mathbb{R}^{h}$, where $h$ depends on $\mu$. The value of $h$ is determined from the CCA, where we approximately assign the mapped objects dimensionality to the number of elements found in the interval $[l_{min},l_{max}]$ for each $\mu$, see Fig.~\ref{fig:2}(b,c). In order to construct those projected states, we use all the delay-coordinates provided by the network, i.e. the full range $[l_{min},l_{max}]$ for each value of $\mu$, as follows  

\begin{eqnarray}
\varphi(\mathbf{y}^{IN}_{n}) = [\varphi_{l_{1}}(\mathbf{y}^{IN}_{n}), \varphi_{l_{2}}(\mathbf{y}^{IN}_{n}), \ldots, \varphi_{l_{h}}(\mathbf{y}^{IN}_{n})], \qquad \\ \varphi(\mathbf{y}^{IN}_{n+1}) = [\varphi_{l_{1}}(\mathbf{y}^{IN}_{n+1}), \varphi_{l_{2}}(\mathbf{y}^{IN}_{n+1}), \ldots, \varphi_{l_{h}}(\mathbf{y}^{IN}_{n+1})];  \qquad
\end{eqnarray}
where $\{  \varphi_{l_{1}}(\mathbf{y}^{IN}_{n}), \varphi_{l_{2}}(\mathbf{y}^{IN}_{n}), \ldots \}$ are node responses lagged at $[l_{min},l_{max}]$. The size of the interval $[l_{min},l_{max}]$ depends on the value of $\mu$, as it was shown by Fig.~\ref{fig:2}(b,c), where we find a broader distribution of delay-coordinates for higher values of $\mu$. 

The \textit{interstate distances} $\| \mathbf{y}^{IN}_{n+1}- \mathbf{y}^{IN}_{n} \|$ and $\| \varphi(\mathbf{y}^{IN}_{n+1})- \varphi(\mathbf{y}^{IN}_{n}) \|$, have to be bounded in the interval $[(1-\epsilon_{1}),(1+\epsilon_{2})]$ according to
\begin{equation}
\dfrac{\| \varphi(\mathbf{y}^{IN}_{n+1})- \varphi(\mathbf{y}^{IN}_{n}) \| }{\| \mathbf{y}^{IN}_{n+1}- \mathbf{y}^{IN}_{n} \|}  \in [(1-\epsilon_{1}),(1+\epsilon_{2})].
\end{equation}
Under these conditions, we can claim that the transformation by the rRNN agrees with a nonlinear random projections. Estimating limits $\{ \epsilon_{1}, \epsilon_{2} \}$ requires to find the inferior $\epsilon_{min}$, and superior $\epsilon_{max}$ interstate distance limits:
\begin{equation}
\dfrac{\| \varphi(\mathbf{y}^{IN}_{n+1})- \varphi(\mathbf{y}^{IN}_{n}) \|_{min} }{\| \mathbf{y}^{IN}_{n+1}- \mathbf{y}^{IN}_{n} \|}  =\epsilon_{min}; 
\end{equation}
\begin{equation}
 \dfrac{\| \varphi(\mathbf{y}^{IN}_{n+1})- \varphi(\mathbf{y}^{IN}_{n}) \|_{max} }{\| \mathbf{y}^{IN}_{n+1}- \mathbf{y}^{IN}_{n} \|} = \epsilon_{max}, \label{ercc1}
\end{equation}
where $\epsilon_{1}$ and $\epsilon_{2}$ are calculated by isolating these constants from $\epsilon_{min}=(1-\epsilon_{1})$, and $\epsilon_{max}=(1+\epsilon_{2})$. These limits contain information about the minimum and maximum distortions that we can find in order to get the best neighbors in the rRNN. $\| \varphi(\mathbf{y}^{IN}_{n+1})- \varphi(\mathbf{y}^{IN}_{n}) \|_{min}$ and $\| \varphi(\mathbf{y}^{IN}_{n+1})- \varphi(\mathbf{y}^{IN}_{n}) \|_{max}$ are calculated by using Euclidean distance under minimum and maximum norms,
\begin{eqnarray}
\| \varphi(\mathbf{y}^{IN}_{n+1})- \varphi(\mathbf{y}^{IN}_{n}) \|_{min} = \qquad \qquad \qquad \qquad \qquad \qquad
\nonumber \\ 
\left(\sum_{l_{g}=l_{min}}^{l_{max}} [\varphi_{l_{g}}(\mathbf{y}^{IN}_{n+1})-\varphi_{l_{g}}(\mathbf{y}^{IN}_{n})]^{2}_{min}\right)^{1/2},   \label{ercc2}	
\end{eqnarray}
\begin{eqnarray}
\| \varphi(\mathbf{y}^{IN}_{n+1})- \varphi(\mathbf{y}^{IN}_{n}) \|_{max}= \qquad \qquad \qquad \qquad \qquad \qquad
\nonumber \\ 
\left(\sum_{l_{g}=l_{min}}^{l_{max}} [\varphi_{l_{g}}(\mathbf{y}^{IN}_{n+1})-\varphi_{l_{g}}(\mathbf{y}^{IN}_{n})]^{2}_{max}\right)^{1/2}, \label{ercc3}	
\end{eqnarray}
where $ \varphi_{l_{g}}(\mathbf{y}^{IN}_{n})$ are node responses lagged at $l_{g} \in [l_{min},l_{max}]$, $\forall g=1,2,\ldots, h$.

Here, we therefore identify the smallest and largest distances $[\varphi_{l_{g}}(\mathbf{y}^{IN}_{n+1})-\varphi_{l_{g}}(\mathbf{y}^{IN}_{n})]_{min, max}$ along each delay coordinate. Then, it is true that these smallest and largest distances bound the Euclidean distance of these.
Finally, we determine $\| \mathbf{y}^{IN}_{n+1}- \mathbf{y}^{IN}_{n} \|$ via
\begin{widetext}
\begin{equation}
\| \mathbf{y}^{IN}_{n+1}- \mathbf{y}^{IN}_{n} \| = \sqrt{(y^{IN}_{n+1}-y^{IN}_{n})^{2}+ \cdots + (y^{IN}_{(n+1)+(M-1)\tau_{0}}-y^{IN}_{n+(M-1)\tau_{0}})^{2}}.	\label{ercd2}
\end{equation}
\end{widetext}

\bibliography{tesis}

\end{document}